\pdfoutput=1

\documentclass[11pt]{article}

\usepackage{ACL2023}

\usepackage{times}
\usepackage{latexsym}

\usepackage[T1]{fontenc}

\usepackage[utf8]{inputenc}

\usepackage{microtype}

\usepackage{inconsolata}

\usepackage{amsfonts} 
\usepackage{amssymb} 
\usepackage{amsmath} 
\usepackage{pifont} 
\usepackage{tabularx} 
\usepackage{arydshln} 
\usepackage{xcolor} 
\usepackage{mathtools} 
\usepackage{adjustbox} 
\usepackage{multirow} 
\usepackage{enumitem}
\usepackage{CJKutf8} 
\usepackage{booktabs} 
\hypersetup{colorlinks=true, citecolor=darkblue, linkcolor=darkblue, urlcolor=darkblue, bookmarksopen=true, bookmarksopenlevel=2, bookmarksnumbered=true}
\usepackage{bbding} 
\usepackage{wasysym}
\usepackage{caption}
\usepackage{subcaption}
\usepackage{tikz}
\usepackage[edges]{forest}
\usepackage{pgfplots}
\usepackage{nicefrac}
\usepackage{verbatim}
\usepackage{bm}
\usepackage[british,american]{babel}
\usepackage{courier}

\usepackage[ruled,linesnumbered]{algorithm2e}
\SetKwInOut{Parameter}{Parameters}

\usepackage{amsthm} 
\theoremstyle{definition}

\newif\ifprint
\printtrue

\newcommand{\cmark}{\ding{51}}
\newcommand{\xmark}{\ding{55}}

\newcommand{\Tref}[1]{Table~\ref{#1}}
\newcommand{\Eref}[1]{Equation~\eqref{#1}}
\newcommand{\Fref}[1]{Figure~\ref{#1}}
\newcommand{\Sref}[1]{Section~\ref{#1}}

\newcommand{\Apref}[1]{Appendix~\ref{#1}}

\usepackage{makecell}
\usepackage{tabulary}
\definecolor{demphcolor}{RGB}{144,144,144}
\newcommand{\demph}[1]{\textcolor{demphcolor}{#1}}
\definecolor{mygray}{gray}{0.4}

\newlength\savewidth\newcommand\shline{\noalign{\global\savewidth\arrayrulewidth
  \global\arrayrulewidth 1pt}\hline\noalign{\global\arrayrulewidth\savewidth}}

\newcommand{\tablestyle}[2]{\setlength{\tabcolsep}{#1}\renewcommand{\arraystretch}{#2}\centering} 

\newcommand{\vqa}[1]{VQAv2}

\newcommand{\method}[0]{{ManagerTower}}
\newcommand{\metername}[0]{{METER}}
\newcommand{\btname}[0]{{BridgeTower}}
\newcommand{\vlmoname}[0]{\textsc{VLMo}}
\newcommand{\twotower}[0]{{Two-Tower}}

\newcommand{\modelbase}[0]{$_{\text{BASE}}$}
\newcommand{\modellarge}[0]{$_{\text{LARGE}}$}
\newcommand{\modelhuge}[0]{$_{\text{HUGE}}$}
\newcommand{\rmean}[0]{R$_\text{MEAN}$}

\newcommand{\eg}[0]{\textit{e.g.},}
\newcommand{\ie}[0]{\textit{i.e.},}
\newcommand{\etc}[0]{\textit{etc}.}
\newcommand{\vs}[0]{\textit{vs}.}

\definecolor{gain}{HTML}{34a853}  %

\definecolor{lost}{HTML}{ea4335}  %

\definecolor{baselinecolor}{gray}{.9}

\title{
	ManagerTower: Aggregating the Insights of Uni-Modal Experts 
	\\ for Vision-Language Representation Learning
}

\author{
	Xiao Xu\textsuperscript{\rm 1, 3}\thanks{\ \ Contribution during internship at Microsoft.},
	Bei Li\textsuperscript{\rm 2, 3},
    Chenfei Wu\textsuperscript{\rm 3},
	Shao-Yen Tseng\textsuperscript{\rm 4},
	Anahita Bhiwandiwalla\textsuperscript{\rm 4}, \\
    \bf{Shachar Rosenman}\textsuperscript{\rm 4},
    \bf{Vasudev Lal}\textsuperscript{\rm 4},
    \bf{Wanxiang Che}\textsuperscript{\rm 1\thanks{\ \ Contact Person}},
    \bf{Nan Duan}\textsuperscript{\rm 3$^\dagger$} \\
	\textsuperscript{\rm 1}Harbin Institute of Technology, Harbin, China,
	\textsuperscript{\rm 2}Northeastern University, Shenyang, China \\
    \textsuperscript{\rm 3}Microsoft Research Asia, \textsuperscript{\rm 4}Intel Labs, Cognitive Computing Research \\
    \texttt{\{xxu,car\}@ir.hit.edu.cn, libei\_neu@outlook.com} \\
	\texttt{\{chenfei.wu,nanduan\}@microsoft.com, shao-yen.tseng@intel.com} \\
	\texttt{\{anahita.bhiwandiwalla,shachar.rosenman,vasudev.lal\}@intel.com} \\
}

\begin{document}
\maketitle
\begin{abstract}
	\twotower{} Vision-Language (VL) models have shown promising improvements on various downstream VL tasks.
	Although the most advanced work improves performance by building bridges between encoders, it suffers from ineffective layer-by-layer utilization of uni-modal representations and
	cannot flexibly exploit different levels of uni-modal semantic knowledge.
	In this work, we propose \method{}, a novel VL model architecture that gathers and combines the insights of pre-trained uni-modal experts at different levels.
	The managers introduced in each cross-modal layer can adaptively aggregate uni-modal semantic knowledge to facilitate more comprehensive cross-modal alignment and fusion.
	\method{} outperforms previous strong baselines both with and without Vision-Language Pre-training (VLP).
	With only $4$M VLP data, \method{} achieves superior performances on various downstream VL tasks, especially $79.15\%$ accuracy on VQAv2 Test-Std, $86.56\%$ IR@1 and $95.64\%$ TR@1 on Flickr30K.
	Code and checkpoints are available at \url{https://github.com/LooperXX/ManagerTower}.
\end{abstract}

\section{Introduction}
\label{sec:introduction}
In recent years, there has been a growing interest in the field of Vision-Language (VL) representation learning due to the development of Vision-Language Pre-training (VLP) techniques.
VLP aims to learn transferable multi-modal knowledge from large-scale image-text pairs, which can further improve the performance of various downstream VL tasks, such as visual question answering~\citep{balanced_vqa_v2}, visual entailment~\citep{xie2019visual}, visual reasoning~\citep{suhr2019corpus}, and image-text retrieval~\citep{young-etal-2014-image}.

\begin{figure}[t]
	\centering
	\includegraphics[width=0.4\textwidth]{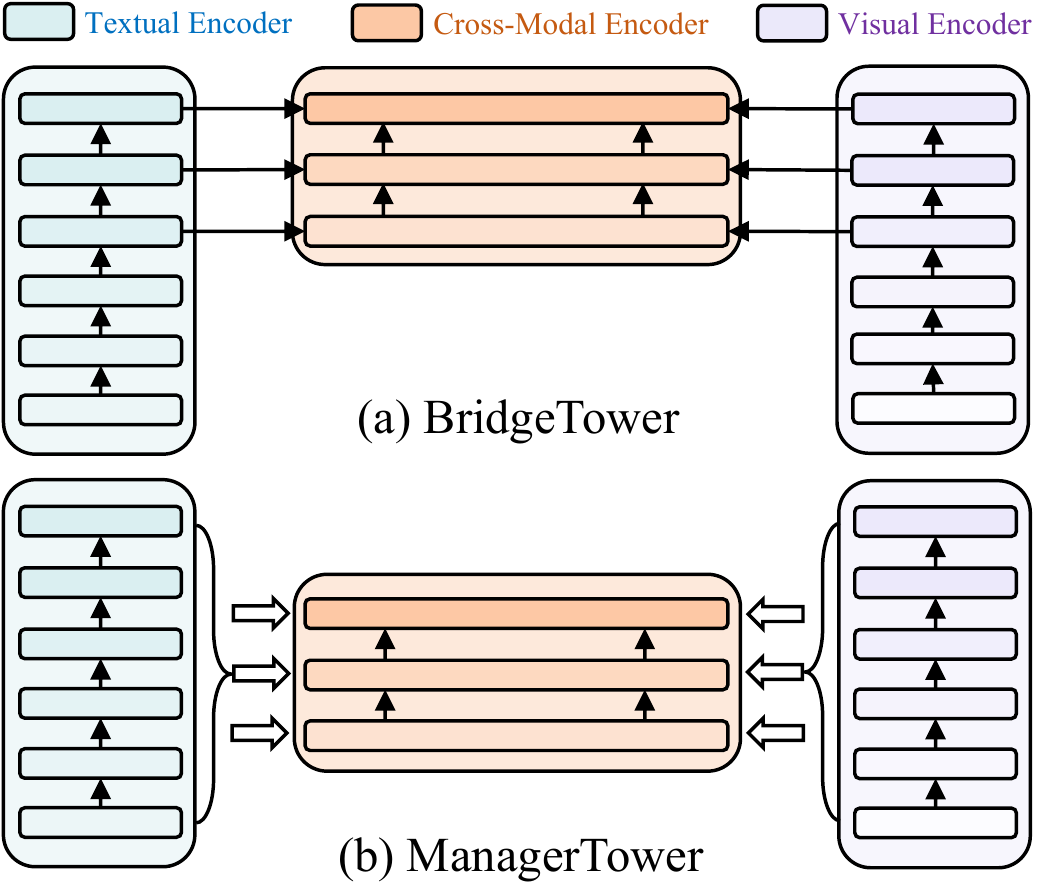}
	\caption{
		Brief illustrations of \btname{} and \method{}.
		Hollow arrows indicate the transmission of multi-layer uni-modal representations in \method{} instead of layer-by-layer transmission in \btname{}.
	}
	\label{fig:overview}
\end{figure}

Visual and textual modalities in VL models are typically processed by uni-modal encoders and subsequently fused in a cross-modal encoder.
This general architecture can be referred to as the \twotower{} architecture.
\metername{}~\citep{dou2021meter} and \btname{}~\citep{xu2022bridge} are two representative \twotower{} VL models.
\metername{} uses CLIP-ViT~\citep{radford2021learning} and RoBERTa~\citep{liu2019roberta} as pre-trained uni-modal encoders, but it ignores different levels of uni-modal semantic knowledge in them and only feeds the last-layer outputs of each uni-modal encoder into the cross-modal encoder.
In an effort to address this issue, as illustrated in~\Fref{fig:overview}(a), \btname{} connects multiple top uni-modal layers with each cross-modal layer in a layer-by-layer fashion to exploit unimodal semantic knowledge at different levels.

In this work, we build upon the research of \btname{} and advance it in two aspects.
Specifically, we address the limitations of \btname{}:
($i$) its layer-by-layer utilization of different uni-modal layer representations is ineffective.
Each cross-modal layer can only utilize an artificially-connected uni-modal layer representation, thus restricting the exploitation of different levels of uni-modal semantic knowledge.
~($ii$) the number of cross-modal layers is tied to the number of uni-modal layer representations it used, thus limiting its scalability and capability. For example, increasing the number of uni-modal layer representations used requires a corresponding increase in the number of cross-modal layers.
This leads to an increase in the number of parameters and computation cost, while does not always result in performance improvements
as demonstrated by~\citet{xu2022bridge}.

As shown in~\Fref{fig:overview}(b), we propose a novel VL model architecture, \method{}, that aggregates multi-layer uni-modal representations via managers in each cross-modal layer.
Each manager takes multi-layer uni-modal representations as the \textbf{insights} of pre-trained uni-modal \textbf{experts} at different levels, and then \textbf{adaptively} aggregates them to facilitate more comprehensive cross-modal alignment and fusion.
More concretely, inspired by the linear combination of layers~\citep{wang-etal-2019-learning-deep} method, we adapt it as the Static Aggregation of Experts (SAE) manager and then remove redundant information to design the Static Aggregation of Uni-modal Experts (SAUE) manager, which focuses on aggregating uni-modal semantic knowledge.
We further propose the Adaptive Aggregation of Uni-modal Experts (AAUE) manager to adaptively aggregate multi-layer uni-modal representations for each token in different cross-modal layers.
Moreover, in principle, managers can be easily integrated into any cross-modal encoders and work well with any uni-modal encoders, making \method{} scalable and flexible.

We first explore the feasibility of various designs of managers by evaluating and analyzing the performance on VQAv2 and Flickr30K datasets.
Then, we pre-train \method{} with commonly used $4$M VLP data and evaluate it on various downstream VL tasks.
With the same pre-training and fine-tuning settings and uni-modal backbones as previous strong baselines such as \metername{} and \btname{},
\method{} achieves superior performances on various downstream VL tasks, especially $79.15\%$ accuracy on VQAv2 Test-Std, $86.56\%$ IR@1 and $95.64\%$ TR@1 on Flickr30K.
It outperforms not only many base-size models pre-trained on $4$M data but also some models pre-trained on more data and/or with larger size.

\section{Preliminary}
\label{sec:preliminary}
In this work, for a fair comparison with \metername{} and \btname{}, we use the same cross-modal encoder and pre-trained uni-modal encoders.

\subsection{Visual Encoder}
CLIP-ViT, the visual encoder of CLIP~\citep{radford2021learning}, has been widely used in VL models~\citep{shen2021much, dou2021meter}.
It reshapes each input image into a flattened patch sequence and prepends a \texttt{[class]} token to the sequence.
After a linear projection, position embeddings are added to the sequence to get the input visual representation $\mathbf{V}_0$.
The $\ell^{\text{\,th}}$ visual layer representation can be computed as: $\mathbf{V}_{\ell} \!=\! \operatorname{Encoder}^\mathrm{V}_\ell(\mathbf{V}_{\ell-1}), \ell \!=\! 1 \dots L_\mathrm{V}$, where $\ell$ is the layer
index and $L_\mathrm{V}$ is the number of layers of the visual encoder.

\begin{figure*}[t]
	\centering
	\includegraphics[width=0.93\textwidth]{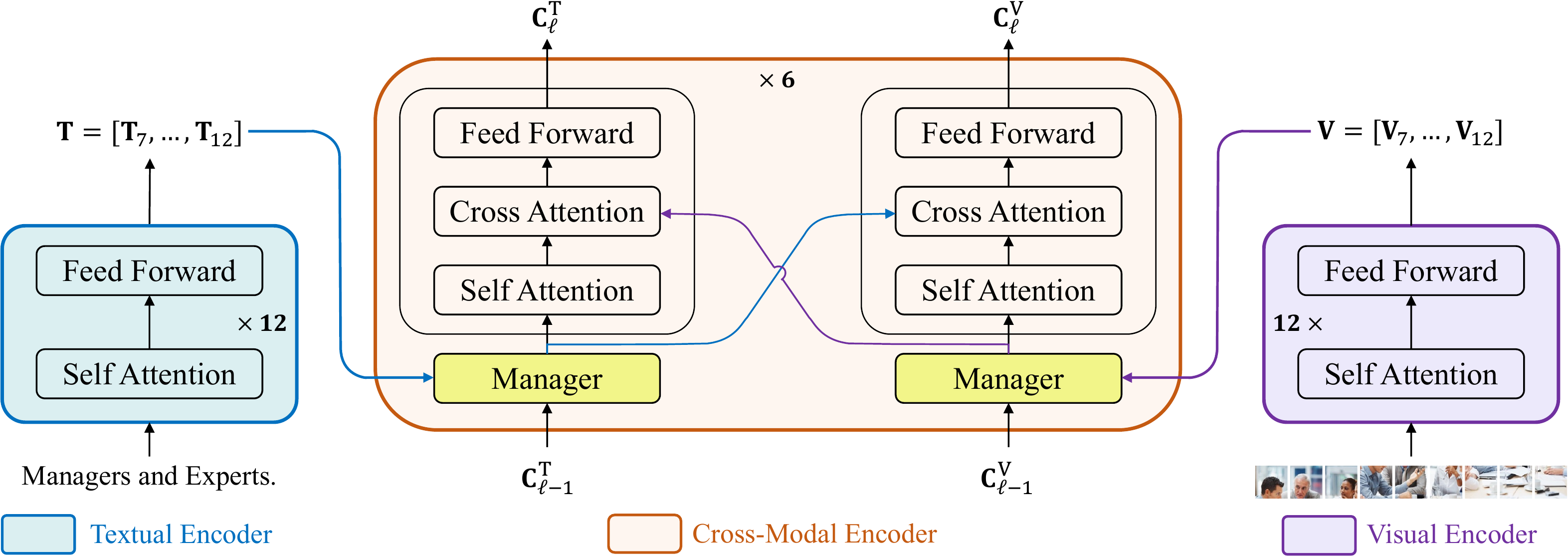}
	\caption{
	An illustration of \method{}, a textual manager and a visual manager are introduced in each cross-modal layer.
	Top $\mathrm{N}\!=\!6$ uni-modal layer representations $\mathbf{T},\mathbf{V}\!\in\!\mathbb{R}^{\mathrm{N} \times \mathrm{L} \times \mathrm{D}}$ and output representations of the previous cross-modal layer ${\mathbf{C}}^\mathrm{T}_{\ell-1}, {\mathbf{C}}^\mathrm{V}_{\ell-1}, \ell \!=\! 1 \dots 6$ are fed into the textual manager $\mathcal{M}_{\ell}^{\mathrm{T}}$ and visual manager $\mathcal{M}_{\ell}^{\mathrm{V}}$, respectively.
	$\mathrm{N}$ is the number of pre-trained uni-modal experts we used, 
	$\mathrm{L}$ is the length of the input sequence.
	}
	\label{fig:framework}
\end{figure*}

\subsection{Textual Encoder}
RoBERTa~\citep{liu2019roberta} is widely used in the field of VL~\citep{dou2021meter,li2022unimo} due to its robust performance.
It tokenizes the input text with the byte-level Byte-Pair Encoding (BPE)~\citep{sennrich-etal-2016-neural,radford2019language} and adds $\texttt{[<s>]}$ and $\texttt{[</s>]}$ tokens to the start and end of the sequence, respectively.
Then, it applies word embeddings and positional embeddings to the tokenized sequence to get the input textual representation $\mathbf{T}_0$.
Similarly, the $\ell^{\text{\,th}}$ textual layer representation can be computed as: $\mathbf{T}_{\ell} \!=\! \operatorname{Encoder}^\mathrm{T}_\ell(\mathbf{T}_{\ell-1}), \ell \!=\! 1 \dots L_\mathrm{T}$, where $L_\mathrm{T}$ is the number of layers of the textual encoder.

\subsection{Cross-Modal Encoder}
\label{sec:cross-modal-encoder}
We adopt the transformer encoder~\citep{vaswani2017attention} with the co-attention mechanism as the cross-modal encoder~\citep{lu2019vilbert}.
For each cross-modal layer, each modality has a multi-head self-attention (MSA) block, a multi-head cross-attention (MCA) block, and a feed-forward (FFN) block.
The MCA block allows the visual part of the cross-modal encoder to attend to the textual part and vice versa.
Each cross-modal layer is denoted as $\operatorname{Encoder}^\mathrm{C}_\ell, \ell \!=\! 1 \dots L_\mathrm{C}$, where $L_\mathrm{C}$ is the number of cross-modal layers.
For brevity, the $\ell^{\text{\,th}}$ cross-modal layer computes as:
\begin{gather}
	\tilde{\mathbf{C}}^\mathrm{V}_{\ell} = {\mathbf{C}}^\mathrm{V}_{\ell-1}, \label{eq:prev1} \\
	\tilde{\mathbf{C}}^\mathrm{T}_{\ell} = {\mathbf{C}}^\mathrm{T}_{\ell-1}, \label{eq:prev2} \\
	\mathbf{C}^\mathrm{V}_{\ell}, \mathbf{C}^\mathrm{T}_{\ell} = \operatorname{Encoder}^\mathrm{C}_\ell(\tilde{\mathbf{C}}^\mathrm{V}_{\ell}, \tilde{\mathbf{C}}^\mathrm{T}_{\ell}),
\end{gather}
where $\mathbf{C}^\mathrm{V}_{\ell}, \mathbf{C}^\mathrm{T}_{\ell}$ are the output representations of the visual and textual part at the $\ell^{\text{\,th}}$ layer, $\tilde{\mathbf{C}}^\mathrm{V}_{\ell}, \tilde{\mathbf{C}}^\mathrm{T}_{\ell}$ are inputs of each part. $\mathbf{C}^\mathrm{V}_{0}, \mathbf{C}^\mathrm{T}_{0}$ are initialized with the last-layer representations from uni-modal encoders: $\mathbf{C}^\mathrm{V}_{0} \!=\! \mathbf{V}_{L_\mathrm{V}}\mathbf{W}_\mathrm{V}, \mathbf{C}^\mathrm{T}_{0} \!=\! \mathbf{T}_{L_\mathrm{T}}\mathbf{W}_\mathrm{T}$, where $\mathbf{W}_\mathrm{V},\mathbf{W}_\mathrm{T}$ are linear cross-modal projections.
In this work, we use the same default setting as \btname{} for a fair comparison: $L_\mathrm{V}\!=\!L_\mathrm{T}\!=\!12,L_\mathrm{C}\!=\!6$, and only top $\mathrm{N}\!=\!6$ uni-modal layer representations are used.

\subsection{Utilization of Uni-Modal Experts}
Different layers of uni-modal encoders encoding different levels of semantic information are well demonstrated in vision~\citep{dosovitskiy2020image,raghu2021vision,naseer2021intriguing} and language~\cite{peters-etal-2018-dissecting,liu2019linguistic,jawahar-etal-2019-bert}.
According to ~\citet{dosovitskiy2020image} and~\citet{raghu2021vision}, lower layers of ViT tend to attend both locally and globally, while higher layers primarily focus on global information.
Similarly,~\citet{jawahar-etal-2019-bert} found that the intermediate layers of BERT~\citep{devlin-etal-2019-bert} encode a hierarchy of linguistic information, with surface features at the bottom, syntactic features in the middle, and semantic features at the top.\looseness=-1

In the field of VL, some works have explored the usage of pre-trained multi-layer uni-modal representations~\citep{dou2021meter,xu2022bridge}.
They simply feed the weighted sum of uni-modal layer representations into the first cross-modal layer, or layer-by-layer exploit multiple top uni-modal layer representations in each cross-modal layer.
In this work, we take each layer of the pre-trained uni-modal encoder as a uni-modal \textbf{expert}, and the output representation of each layer as the \textbf{insight} of the uni-modal expert into the current input.

\section{Manager Design}
\label{sec:manager-design}
\Fref{fig:framework} depicts the overall framework of \method{}.
It introduces managers in each cross-modal layer to adaptively aggregate the insights of pre-trained uni-modal experts at different levels.
In the subsequent subsections, we will elaborate on the detailed design schema for the three types of managers, and conclude with the cross-modal encoder with our well-designed managers.\footnote{More details on pre-training objectives and downstream fine-tuning are described in~\Apref{appendix:implementation_details}.}

\subsection{Static Aggregation of Experts}
\label{sec:sae}
The effectiveness of layer fusion in learning comprehensive representations has been well demonstrated in machine translation~\citep{wang-etal-2018-multi-layer,wang-etal-2019-learning-deep,wei-etal-2020-multiscale}.
Motivated by this, we decide to apply this technique in the context of VL.
As a preliminary approach, we choose to utilize the linear combination of layers method~\citep{wang-etal-2019-learning-deep}, which is a simple yet effective way to aggregate the representations of previous layers through the use of learned weights in each encoder layer.

A natural idea is to adapt it to aggregate uni-modal and cross-modal output representations of all previous layers.
We name it the Static Aggregation of Experts (SAE) manager.
The calculation of the $\ell^{\text{\,th}}$ visual manager is:
\begin{align}
	\label{eq:sae}
	 & \mathcal{M}_{\ell}^{\mathrm{V}}(\mathbf{V}_{7}, \dots, \mathbf{V}_{12}, \mathbf{C}^\mathrm{V}_{1}, \dots, \mathbf{C}^\mathrm{V}_{\ell-1}) =                                                                              \\
	 &  \sum_{i=1}^{\ell-1}{\mathbf{W}_{i+6}^{\mathrm{V}, \ell} \odot \operatorname{LN}(\mathbf{C}^\mathrm{V}_{i})} \!+\! \sum_{i=1}^{6}{\mathbf{W}_{i}^{\mathrm{V}, \ell} \odot \operatorname{LN}(\mathbf{V}_{i+6})}, \nonumber
\end{align}
where $\mathcal{M}_{\ell}^{\mathrm{V}}$ denotes the manager for the visual part of the $\ell^{\text{\,th}}$ cross-modal layer, $\mathbf{W}^{\mathrm{V}, \ell} \!\in\! \mathbb{R}^{(6+\ell-1) \times \mathrm{D}}$ is a learnable parameter matrix, $\odot$ denotes the element-wise product operation and $\operatorname{LN}(\cdot)$ denotes Layer Normalization~\citep{ba2016layer}.
The $\operatorname{softmax}$ with a learnable temperature is used to normalize $\mathbf{W}^{\mathrm{V}, \ell}$.
We then omit the superscript $^{\mathrm{V}, \ell}$ of $\mathbf{W}$ for brevity.
The learned aggregation weight $\mathbf{W}$ is initialized with $\frac{1}{6+\ell-1}$ on average in order to assign equal weights to the output representation of all previous layers.

However, directly applying SAE to VL models is non-trivial, since it does not bring a desired performance improvement compared to \btname{} but led to a significant performance decrease.
We posit that this decrease may be due to the average initialization of $\mathbf{W}$ not being suitable for cross-modal and pre-trained uni-modal output representations as they have different scales.
To investigate this hypothesis, we propose dividing the parameter matrix $\mathbf{W}$ into uni-modal and cross-modal parts and initializing them with $\frac{1}{6}$ and $\frac{1}{\ell-1}$, respectively,\footnote{
	We also try some different initialization methods: one, progressive, exponential moving average, \btname{}-like, \etc{}, but the results are similar to or lower than the average.
} and also learn the $\operatorname{softmax}$ temperature separately.
The experimental result yield a significant improvement compared to the direct application of SAE, but a limited improvement compared to \btname{}.
These observations provide a compelling argument for re-examining how to aggregate multi-layer pre-trained uni-modal representations.

\subsection{Static Aggregation of Uni-Modal Experts}
\label{sec:saue}
Since~\Eref{eq:sae} can be divided into uni-modal and cross-modal parts, by computing the cosine similarity of aggregated uni-modal/cross-modal representations between every two consecutive textual/visual managers,
we further analyze the insights aggregated by different SAE managers.

\begin{figure}[t]
	\centering
	\includegraphics[width=0.48\textwidth]{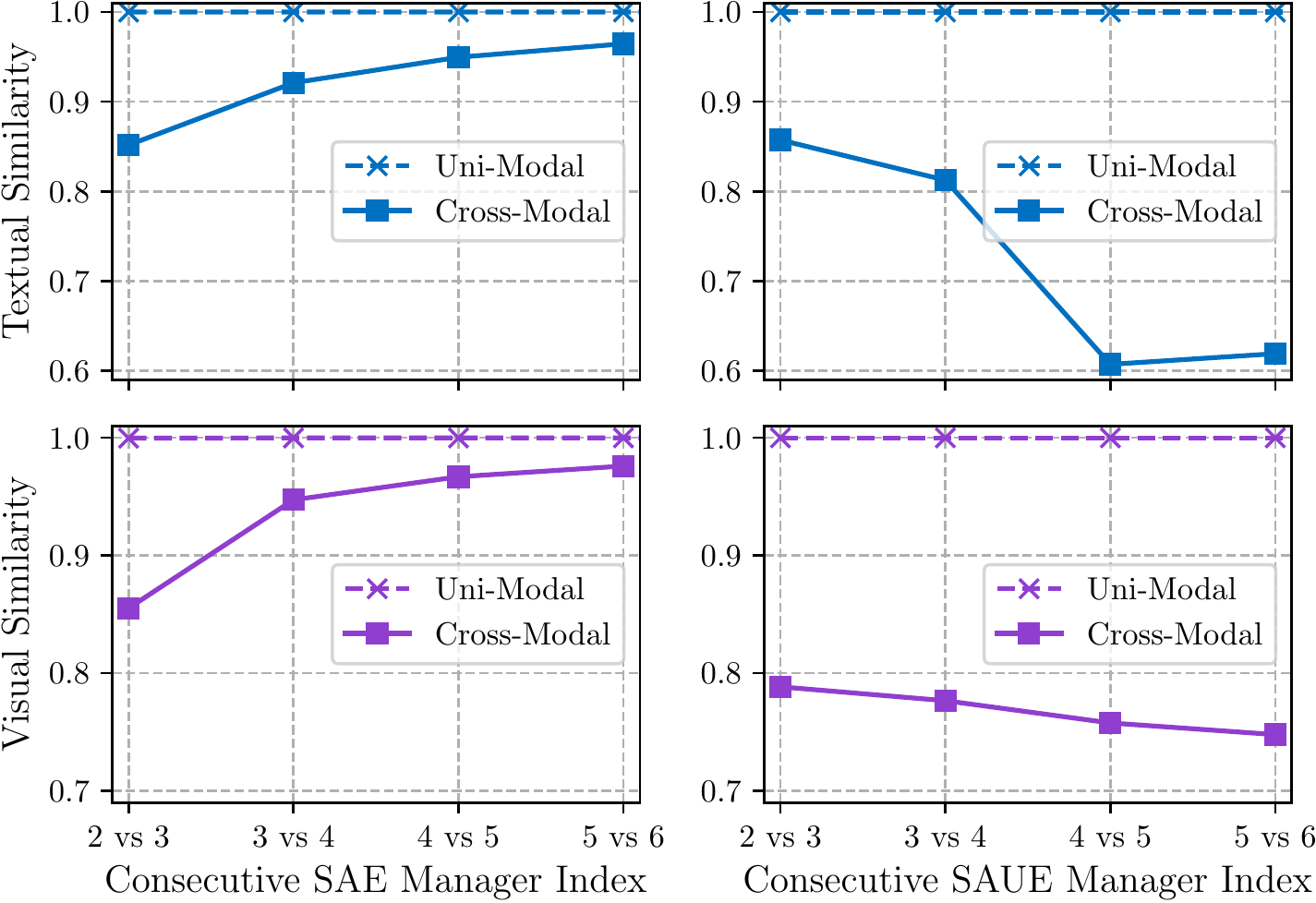}
	\caption{
		Cosine similarity of aggregated uni-modal/cross-modal representations between every two consecutive textual/visual managers.
	}
	\label{fig:cosine_similarity_cross_uni}
\end{figure}

As shown in~\Fref{fig:cosine_similarity_cross_uni}, for SAE managers, the uni-modal similarity is always similar to $1$, while the cross-modal similarity increases with depth and gets closer to $1$. 
This indicates that, the uni-modal representations aggregated by different SAE managers are almost identical, and the aggregated cross-modal representations get similar with depth.

We hypothesize that, since different SAE managers provide similar aggregated uni-modal representations to each cross-modal layer, output representation of more preceding cross-modal layers may bring redundant information to confuse the managers.
This leads to aggregated cross-modal representations converging to indistinguishable vectors as the depth increases.

Hence, we propose focusing on aggregating the insights of pre-trained uni-modal experts and keeping only the output representation of the previous cross-modal layer.
We name it the Static Aggregation of Uni-modal Experts (SAUE) manager.
The calculation of the $\ell^{\text{\,th}}$ visual manager becomes:
\begin{equation}
	\begin{split}
		& \mathcal{M}_{\ell}^{\mathrm{V}}(\mathbf{V}_{7}, \dots, \mathbf{V}_{12}, \mathbf{C}^\mathrm{V}_{\ell-1}) = \\
		& {\mathbf{W}_{\mathrm{C}} \odot \operatorname{LN}(\mathbf{C}^\mathrm{V}_{\ell-1})} \!+\! \sum_{i=1}^{6}{\mathbf{W}_{i} \odot \operatorname{LN}\left(\mathbf{V}_{i+6}\right)}, \label{eq:saue}
	\end{split}
\end{equation}
where $\mathbf{W}\!\in\!\mathbb{R}^{6 \times \mathrm{D}}$ and $\mathbf{W}_{\mathrm{C}}\!\in\!\mathbb{R}^{1 \times \mathrm{D}}$ are learnable parameter matrices and initialized with $\frac{1}{6}$ and $1$ on average, respectively.
The $\operatorname{softmax}$ with a learnable temperature only normalizes $\mathbf{W}$.

The significant improvement compared to \btname{} empirically support our hypothesis.
Moreover, in~\Fref{fig:cosine_similarity_cross_uni}, the cross-modal similarity of SAUE decreases with depth, which indicates that comprehensive and distinguishable cross-modal representations are learned as depth increases.

\subsection{Adaptive Aggregation of Uni-Modal Experts}
Although the SAUE manager achieves a significant performance improvement, it still has two limitations:
($i$) $\mathbf{W}$, the learned aggregation weight of uni-modal expert insights, is almost identical between managers in different cross-modal layers, as shown in~\Fref{fig:cosine_similarity_cross_uni}~\&~\ref{fig:aggregation-weights-saue}, 
which is inconsistent with the intuition that the need for uni-modal semantic knowledge varies among cross-modal layers; 
($ii$) in the inference phase, managers in different cross-modal layers use the same aggregation weight of uni-modal expert insights for all tokens in different samples, which does not match the intuition that the need for uni-modal semantic knowledge varies among tokens and samples.

\begin{figure*}[t]
	\centering
	\includegraphics[width=0.9\textwidth]{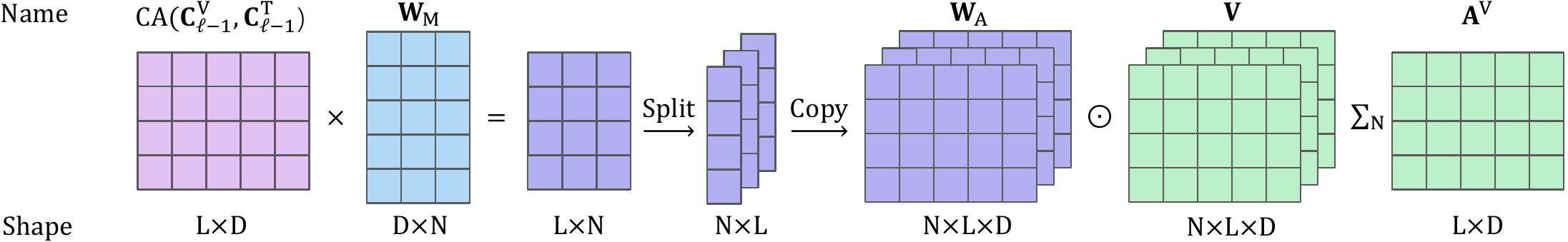}
	\caption{
	An illustration of the calculation of aggregated uni-modal representations $\mathbf{A}^\mathrm{V}\!\in\!\mathbb{R}^{\mathrm{L} \times \mathrm{D}}$ in the visual AAUE manager.
	CA denotes the cross-attention mechanism.
	$\mathrm{N}\!=\!6$.
	We omit $\operatorname{LN}$ and $\operatorname{softmax}$ for brevity.
	}
	\label{fig:details}
\end{figure*}

To address the above limitations, we propose the Adaptive Aggregation of Uni-Modal Experts (AAUE) manager.
During training and inference phases, AAUE managers can adaptively exploit different levels of uni-modal semantic knowledge from pre-trained uni-modal experts, for different tokens in different samples.
Take the visual AAUE manager for example, the calculation of the $\ell^{\text{\,th}}$ visual manager becomes:
\begin{align}
	& \mathcal{M}_{\ell}^{\mathrm{V}}(\mathbf{V}_{7}, \dots, \mathbf{V}_{12}, \mathbf{C}^\mathrm{V}_{\ell-1}) \!=\! \nonumber \\
	& {\mathbf{W}_{\mathrm{C}}\!\odot\!\operatorname{LN}(\mathbf{C}^\mathrm{V}_{\ell-1})}\!+\!\sum_{i=1}^{6}{\mathbf{W}_{\mathrm{A}, i} \odot \operatorname{LN}\left(\mathbf{V}_{i+6}\right)}, \label{eq:aaue} \\
	& \mathbf{W}_\mathrm{A}\!=\!\operatorname{softmax}(\operatorname{LN}(\mathbf{C}^\mathrm{V}_{\ell-\!1})\!\times\!\mathbf{W}_{\mathrm{M}}\!+\!\epsilon), \label{eq:moe}
\end{align}
where $\mathbf{W}_{\mathrm{M}}\!\in\!\mathbb{R}^{\mathrm{D} \times 6}$ is a linear projection layer.
The generated aggregation weights $\mathbf{W}_\mathrm{A}\!\in\!\mathbb{R}^{6 \times \mathrm{L} \times \mathrm{D}}$ can adaptively aggregate uni-modal representations of each token from different levels of pre-trained uni-modal experts.
The $\operatorname{softmax}$ has a learnable temperature and $\epsilon \sim \mathcal N(0, \frac{1}{6^2})$ is a Gaussian noise for exploration of aggregation~\citep{xue2022go}.

Furthermore, to better help managers to exploit uni-modal semantic knowledge for the current cross-modal layer, 
we propose replacing the visual query $\mathbf{C}^\mathrm{V}_{\ell-1}$ in~\Eref{eq:moe} with the cross-modal fused query $\operatorname{CA}(\mathbf{C}^\mathrm{V}_{\ell-1}, \mathbf{C}^\mathrm{T}_{\ell-1})$ to further improve performance, 
where $\operatorname{CA}$ is a cross-attention mechanism.
We visualize $\mathbf{W}_\mathrm{A}$ in \Sref{sec:visualization}.

\subsection{Cross-Modal Encoder with Managers}

Since the $1^{\text{st}}$ cross-modal layer lacks the output representations of the previous cross-modal layer as the query,
we introduce the SAUE managers in the $1^{\text{st}}$ cross-modal layer and the AAUE managers in the subsequent cross-modal layers.
Hence,~\Eref{eq:prev1}~\&~(\ref{eq:prev2}) of the $1^{\text{st}}$ cross-modal layer with SAUE managers becomes:
\begin{gather}
	\tilde{\mathbf{C}}^\mathrm{V}_{1} = \mathcal{M}_{1}^{\mathrm{V}}(\mathbf{V}_{7}, \dots, \mathbf{V}_{12}), \\
	\tilde{\mathbf{C}}^\mathrm{T}_{1} = \mathcal{M}_{1}^{\mathrm{T}}(\mathbf{T}_{7}, \dots, \mathbf{T}_{12}).
\end{gather}

For the $2^{\text{nd}}$ and subsequent cross-modal layers with AAUE managers:
\begin{gather}
	\tilde{\mathbf{C}}^\mathrm{V}_{\ell} = \mathcal{M}_{\ell}^{\mathrm{V}}(\mathbf{V}_{7}, \dots, \mathbf{V}_{12}, \mathbf{C}^\mathrm{V}_{\ell-1}, \mathbf{C}^\mathrm{T}_{\ell-1}), \\
	\tilde{\mathbf{C}}^\mathrm{T}_{\ell} = \mathcal{M}_{\ell}^{\mathrm{T}}(\mathbf{T}_{7}, \dots, \mathbf{T}_{12}, \mathbf{C}^\mathrm{T}_{\ell-1}, \mathbf{C}^\mathrm{V}_{\ell-1}),
\end{gather}
where we omit the modality type and layer index embeddings added to uni-modal layer representations $\mathbf{V}, \mathbf{T}$ in the above equations for simplicity.

\Fref{fig:details} shows adaptive aggregation of the insights of pre-trained visual experts in AAUE manages, 
which is the uni-modal (right) part of Equation~(\ref{eq:aaue}).
As for SAUE managers, they directly broadcast the learned weights $\mathbf{W}\!\in\!\mathbb{R}^{\mathrm{6} \times \mathrm{D}}$ to $\mathbf{W}_\mathrm{A}$ and then aggregate the insights.

\section{Experiments}
\label{sec:experiments}

\subsection{Implementation Details}
\method{} consists of a pre-trained textual encoder, RoBERTa\modelbase{} with $124$M parameters, a pre-trained visual encoder, CLIP-ViT B-224/16 with $86$M parameters, and a randomly-initialized $6$-layer cross-modal encoder with managers which has $113$M$+12$M parameters.
The detailed setting of the cross-modal encoder is the same as \btname{}.
The maximum length of the text sequence is set to $50$, and the image patch size is $16 \times 16$.
We use an image resolution of $384 \times 384$ for Flickr30K and $576 \times 576$ for VQAv2 for a fair comparison with \btname{}.
AdamW~\citep{loshchilov2018decoupled} optimizer with a base learning rate of $2e^{-5}$ and warmup ratio of $0.1$ is used.

\subsection{Investigation and Analysis}
\label{sec:investigation-and-analysis}
In this section, we investigate various designs of managers and evaluate the performance by directly fine-tuning on VQAv2 and Flickr30K without VLP.
Experimental settings are the same as \btname{} for a fair comparison.
Note that uni-modal encoders are initialized with their pre-trained weights.

\subsubsection{Type of Manager}
We first investigate the performance of different types of managers and different queries.
Take the visual manager for example,
based on the top $\mathrm{N}\!=\!6$ visual layer representations $\mathbf{V}\!\in\!\mathbb{R}^{\mathrm{N} \times \mathrm{L} \times \mathrm{D}}$ from CLIP-ViT,
different managers provide the aggregation weights that can be broadcast to $\mathbf{W}_\mathrm{A}$ for aggregating the insights of pre-trained visual experts.
From the perspective of aggregation weights $\mathbf{W}_\mathrm{A}$, the SAE and SAUE managers are \textbf{static} sentence-level managers that share the same aggregation weights for all tokens in different samples.
Correspondingly, the AAUE manager is an \textbf{adaptive} token-level manager that adaptively \textbf{generates} different aggregation weights for different tokens in different samples.
Besides, we also implement \Eref{eq:moe} with commonly used cross-attention and concat-attention mechanisms for comparison.

Results are shown in~\Tref{tab:type-and-query}.
By focusing on aggregating the insights of pre-trained uni-modal experts, the SAUE manager outperforms the SAE manager on both datasets.
Furthermore, with the help of the cross-modal fused query, the AAUE manager achieves substantially better performance than other managers.
This demonstrates the effectiveness of adaptive token-level aggregation with the cross-modal fused query compared to static sentence-level aggregation.
Notably, the cross-modal fused query incorporates output representations of both visual and textual parts of the previous cross-modal layer, which can better help managers to correctly aggregate uni-modal semantic knowledge required by the current cross-modal layer.

\begin{table}[t]
	\tablestyle{2.5pt}{1.0}
	\adjustbox{width=\linewidth}{
		\begin{tabular}{l|c|c|c|c}
			Type                  & Visual Query                                                                & Weight                                           & Test-Dev       & \rmean{}   \\
			\shline
			\demph{BT}  & \demph{-}                                                                           & \demph{$\mathrm{N} \times 1$}                            & \demph{75.91}          & \demph{93.33}      \\
			\hline
			\multirow{2}{*}{SAE}  & -                                                                           & $\mathrm{N} \times 1$                            & 76.19          & 93.57      \\
			                      & -                                                                           & $\mathrm{N} \times \mathrm{D}$                   & 76.18          & 93.73      \\
			\hline
			\multirow{2}{*}{SAUE} & -                                                                           & $\mathrm{N} \times 1$                            & 76.38          & 93.75      \\
			\textbf{}             & -                                                                           & $\mathrm{N} \times \mathrm{D}$                   & 76.55          & 93.82      \\
			\hline
			\multirow{2}{*}{AAUE} & $\mathbf{C}^\mathrm{V}_{\ell-1}$                                            & $\mathrm{N} \times \mathrm{L}$                   & 76.52          & 93.84      \\
			                      & $\mathbf{C}^\mathrm{V}_{\ell-1}, \mathbf{C}^\mathrm{T}_{\ell-1}$            & $\mathrm{N} \times \mathrm{L}$                   & \textbf{76.65} & \bf{93.97} \\
			\hline
			Concat-               & $\mathbf{V}, \mathbf{C}^\mathrm{V}_{\ell-1}$                                & $\mathrm{N} \times \mathrm{L} \times \mathrm{D}$ & 76.38          & 93.78      \\
			Attention             & $\mathbf{V},\mathbf{C}^\mathrm{V}_{\ell-1}, \mathbf{C}^\mathrm{T}_{\ell-1}$ & $\mathrm{N} \times \mathrm{L} \times \mathrm{D}$ & 76.43          & 93.83      \\
			\hline
			Cross-                & $\mathbf{C}^\mathrm{V}_{\ell-1}$                                            & $\mathrm{N} \times \mathrm{L}$                   & 76.41          & 92.15      \\
			Attention             & $\mathbf{C}^\mathrm{V}_{\ell-1}, \mathbf{C}^\mathrm{T}_{\ell-1}$            & $\mathrm{N} \times \mathrm{L}$                   & 76.45          & 92.61
		\end{tabular}
	}
	\caption{
		Performance of different types of managers and different queries on VQAv2 and Flickr30K.
		\rmean{} indicates the mean recall metrics for image-text retrieval.
	}
	\label{tab:type-and-query}
\end{table}

\begin{table}[t]
	\tablestyle{5pt}{1.0}
	\adjustbox{width=\linewidth}{
		\begin{tabular}{c|cc|cc}
			\multirow{2}{*}{$L_\mathrm{C}$} & \multicolumn{2}{c|}{VQAv2 Test-Dev} & \multicolumn{2}{c}{Flickr30K \rmean{}}                                             \\
			                                & BT                                  & \multicolumn{1}{c|}{Ours}              & BT    & \multicolumn{1}{c}{Ours}          \\
			\shline
			2                               & 74.86                               & 75.47 ($\uparrow$\,0.61)               & 92.45 & 93.31 ($\uparrow$\,0.86)          \\
			3                               & 75.33                               & 76.04 ($\uparrow$\,0.71)               & 92.50 & 93.41 ($\uparrow$\,0.91)          \\
			4                               & 75.74                               & 76.26 ($\uparrow$\,0.52)               & 92.76 & 93.59 ($\uparrow$\,0.83)          \\
			6                               & 75.91                               & \textbf{76.65} ($\uparrow$\,0.74)      & 93.33 & \textbf{93.97} ($\uparrow$\,0.64) \\
			8                               & 75.89                               & 76.47 ($\uparrow$\,0.58)               & 93.03 & 93.65 ($\uparrow$\,0.62)          \\
		\end{tabular}
	}
	\caption{
		Performance of \btname{} (BT) and \method{} with different number of cross-modal layers.
	}
	\label{tab:number-of-cross-modal-layers}
\end{table}

\subsubsection{Number of Cross-Modal Layers}
We compare \method{} to \btname{} with different numbers of cross-modal layers in~\Tref{tab:number-of-cross-modal-layers} to further evaluate the effectiveness of \method{}.
Regardless of the number of cross-modal layers, \method{} consistently and significantly outperforms \btname{} on both datasets.

More interestingly, the performance of \method{} with $L_\mathrm{C}\!=\!3$ ($76.04$) is even better than that of \btname{} with $L_\mathrm{C}\!=\!6$ ($75.91$).
Unlike \btname{}, the number of uni-modal layer representations used $\mathrm{N}$ in \method{} is not tied to the number of cross-modal layers $L_\mathrm{C}$ and can be flexibly adjusted.
We fix $\mathrm{N}\!=\!6$ as the default setting.
Therefore, \method{} actually uses the same number of uni-modal layer representations as \btname{}, but achieves even better performance using half the number of cross-modal layers.
This further demonstrates the flexibility and effectiveness of \method{} to adaptively aggregate uni-modal semantic knowledge, compared to layer-by-layer exploitation in \btname{}.

\begin{figure}[t]
	\centering
	\includegraphics[width=0.48\textwidth]{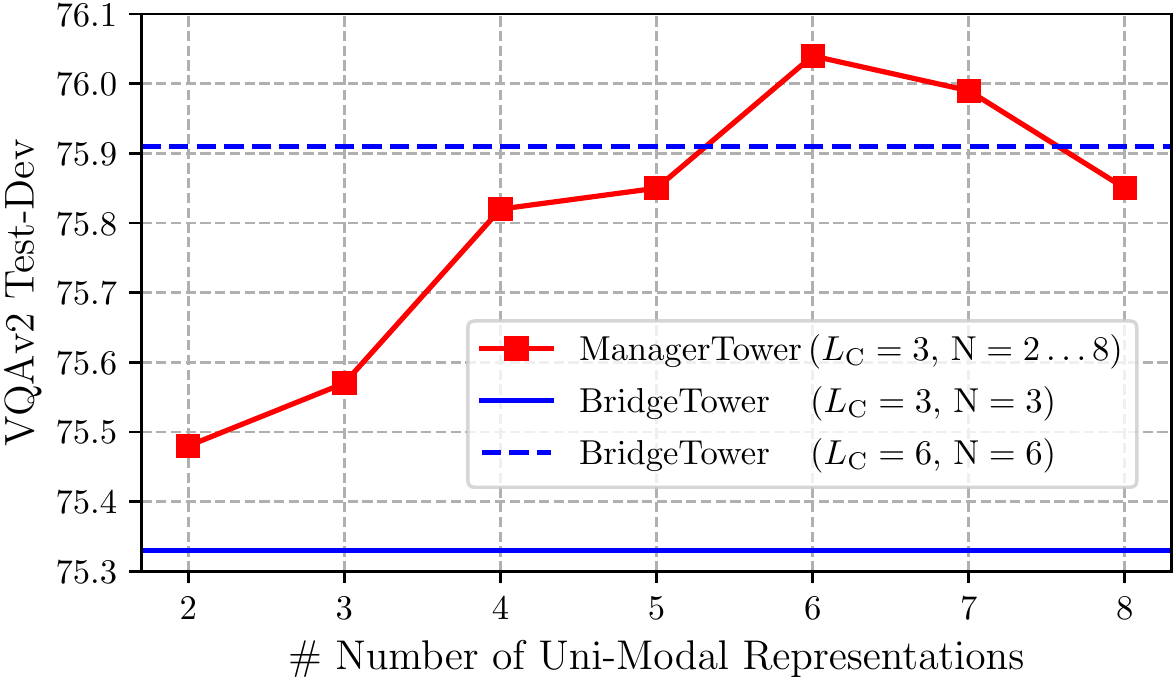}
	\caption{
		Effect of using different numbers of uni-modal representations in \method{}($L_\mathrm{C}\!=\!3, \mathrm{N}=2 \dots 8$).
	}
	\label{fig:number-of-uni-modal-representations}
\end{figure}

\begin{table*}[ht]
	\tablestyle{4.5pt}{1.0} 
	\adjustbox{width=\linewidth}{
		\begin{tabular}{lr|cc|cc|cc|cc}
			\multirow{2}{*}{Model}                             & {\#~Pre-train} & \multicolumn{2}{c|}{VQAv2} & \multicolumn{2}{c|}{SNLI-VE} & \multicolumn{2}{c|}{NLVR$^2$} & \multicolumn{2}{c}{Flickr30K}                                                     \\
			                                                   & Images~        & Test-Dev                   & Test-Std                     & Dev                           & Test                          & Dev        & Test-P     & IR@1       & TR@1       \\
			\shline
			\multicolumn{10}{l}{ { \it{Base-size models pre-trained on 4M public data} } }                                                                                                                                                                      \\
			\hline
			ViLT\modelbase{}~\citep{kim2021vilt}               & 4M             & 71.26                      & -                            & -                             & -                             & 75.70      & 76.13      & 64.4       & 83.5       \\
			UNITER\modelbase{}~\citep{chen2020uniter}\,$\ast$  & 4M             & 72.70                      & 72.91                        & 78.59                         & 78.28                         & 77.18      & 77.85      & 72.52      & 85.90      \\
			UNIMO\modelbase{}~\citep{li2021unimo}              & 4M             & 73.79                      & 74.02                        & 80.00                         & 79.10                         & -          & -          & 74.66      & 89.70      \\
			ALBEF\modelbase{}~\citep{li2021align}\,$\ast$      & 4M             & 74.54                      & 74.70                        & 80.14                         & 80.30                         & 80.24      & 80.50      & 82.8       & 94.3       \\
			\metername{}-Swin\modelbase{}\citep{dou2021meter}  & 4M             & 76.43                      & 76.42                        & 80.61                         & 80.45                         & 82.23      & 82.47      & 79.02      & 92.40      \\
			\vlmoname{}\modelbase{}~\citep{wang2021vlmo}       & 4M             & 76.64                      & 76.89                        & -                             & -                             & 82.77      & 83.34      & 79.3       & 92.3       \\
			\metername{}-CLIP\modelbase{}~\citep{dou2021meter} & 4M             & 77.68                      & 77.64                        & 80.86                         & 81.19                         & 82.33      & 83.05      & 82.22      & 94.30      \\
			\btname{}\modelbase{}~\citep{xu2022bridge}         & 4M             & 78.66                      & 78.73                        & 81.11                         & 81.19                         & 81.85      & 83.09      & 85.83      & 94.73      \\
			\method{}\modelbase{}~(\bf{Ours})                  & 4M             & \bf{79.39}                 & \bf{79.15}                   & \bf{81.26}                    & \bf{81.44}                    & \bf{82.81} & \bf{83.34} & \bf{86.56} & \bf{95.64} \\
			\hline
			\multicolumn{10}{l}{ { \it{Models pre-trained on more data and/or with larger size}}}                                                                                                                                                               \\
			\hline
			UNITER\modellarge{}~\citep{chen2020uniter}\,$\ast$ & 4M             & 73.82                      & 74.02                        & 79.39                         & 79.38                         & 79.12      & 79.98      & 75.56      & 87.30      \\
			UNIMO\modellarge{}~\citep{li2021unimo}             & 4M             & 75.06                      & 75.27                        & 81.11                         & 80.63                         & -          & -          & 78.04      & 89.40      \\
			ALBEF\modelbase{}~\citep{li2021align}\,$\ast$      & 14M            & 75.84                      & 76.04                        & 80.80                         & 80.91                         & 82.55      & 83.14      & 85.6       & 95.9       \\
			SimVLM\modelbase{}~\citep{wang2021simvlm}          & 1.8B           & 77.87                      & 78.14                        & 84.20                         & 84.15                         & 81.72      & 81.77      & -          & -          \\
			BLIP\modelbase{}~\citep{li2022blip}\,$\ast$        & 129M           & 78.24                      & 78.17                        & -                             & -                             & 82.48      & 83.08      & 87.3       & 97.3       \\
			SimVLM\modellarge{}~\citep{wang2021simvlm}         & 1.8B           & 79.32                      & 79.56                        & 85.68                         & 85.62                         & 84.13      & 84.84      & -          & -          \\
		\end{tabular}
	}
	\caption{
		Comparisons with previous models on downstream VL tasks.
		The best score is bolded. 
		$\ast$ indicates that the model also uses VG-QA data to fine-tune on VQAv2.
	}
	\label{tab:main-results}
\end{table*}

\subsubsection{Number of Uni-Modal Experts.}
We further investigate the effect of varying $\mathrm{N}$ in \method{} with $L_\mathrm{C}\!=\!3$.
As shown in~\Fref{fig:number-of-uni-modal-representations}, there exist two interesting observations:
($i$) \method{} ($L_\mathrm{C}\!=\!3, \mathrm{N}\!=\!3$) is still better than \btname{} ($L_\mathrm{C}\!=\!3, \mathrm{N}\!=\!3$).
This indicates that when the same number of uni-modal layer representations are introduced, \method{} allows more effective aggregation of uni-modal semantic knowledge, thus facilitating cross-modal alignment and fusion in each cross-modal layer. ($ii$) the performance of \method{} first increases gradually, but decreases after $\mathrm{N}\!>\!6$.
We assume that lower-layer uni-modal representations may not help \method{} learn cross-modal fusion and also increases the computational cost, which is also consistent with the observation in~\citet{xu2022bridge}.

\subsection{Comparison with Previous Arts}
\paragraph{Pre-train Settings.}
We pre-train \method{} with two standard VLP objectives, masked language modeling (MLM) and image-text matching (ITM), on the commonly used $4$M public data: Conceptual Captions (CC)~\citep{sharma-etal-2018-conceptual}, SBU Captions~\citep{NIPS2011_5dd9db5e}, MSCOCO Captions~\citep{chen2015microsoft}, and Visual Genome (VG)~\citep{krishna2017visual}.
The pre-train settings are the same as \btname{} and \metername{} for a fair comparison.
\method{} is pre-trained for $100$k steps with a batch size of $4096$ and a learning rate of $1e^{-5}$. The image resolution for VLP is $288 \times 288$ and only center-crop~\citep{radford2021learning} is used without any data augmentation.

\begin{figure*}[t]
	\centering
	\includegraphics[width=\textwidth]{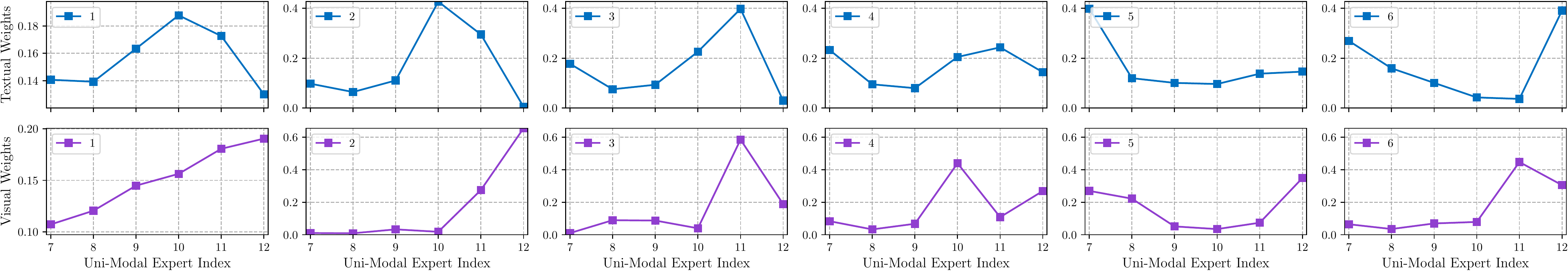}
	\caption{
		A visualization of aggregation weights of textual and visual AAUE managers in each cross-modal layer after VLP.
		The X-axis is the index of the uni-modal expert, and the legend shows the index of the cross-modal layer. 
	}
	\label{fig:aggregation-weights-pre-trained-aaue}
\end{figure*}

\begin{figure*}[t]
	\centering
	\includegraphics[width=\textwidth]{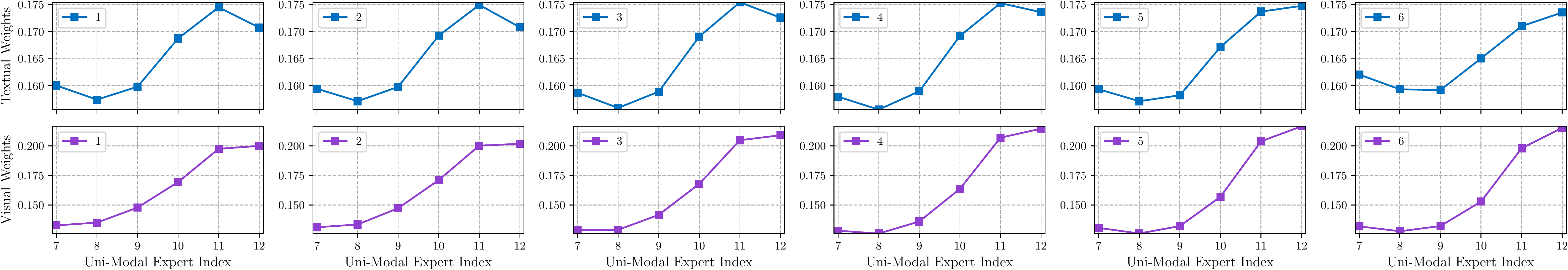}
	\caption{
		A visualization of aggregation weights of textual and visual SAUE managers in each cross-modal layer.
		The X-axis is the index of the uni-modal expert, and the legend shows the index of the cross-modal layer.\looseness=-1
	}
	\label{fig:aggregation-weights-saue}
\end{figure*}

\paragraph{Main Results.}
\Tref{tab:main-results} shows the performance of \method{} compared with other previous works on various downstream VL tasks.
\method{} achieves superior performances on these datasets with only $4$M VLP data.
With the same pre-training and fine-tuning settings and uni-modal backbones as previous strong baselines \metername{} and \btname{},
\method{} significantly improves performances on various downstream VL tasks, especially $79.15\%$ accuracy on VQAv2 Test-Std, $86.56\%$ IR@1 and $95.64\%$ TR@1 on Flickr30K.
This further demonstrates that with all other factors fixed, compared to \btname{} that introduces bridges to \metername{}, \method{} allows more effective aggregation of multi-layer uni-modal representations via well-designed managers.
Managers can adaptively aggregate more accurate uni-modal semantic knowledge to facilitate comprehensive cross-modal alignment and fusion in each cross-modal layer.
Notably, \method{} not only outperforms many base-size models pre-trained on $4$M data, but also surpasses some models pre-trained on more data and/or with larger size.

\subsection{Visualization of Aggregation Weights}
\label{sec:visualization}
We delve into managers by visualizing the average aggregation weights they generate for each cross-modal layer over all samples in VQAv2 Valid in~\Fref{fig:aggregation-weights-pre-trained-aaue}.
For each row, the first column shows the learned aggregation weights  of SAUE managers.
The other five columns show the aggregation weights generated by AAUE managers and share the Y-axis to provide easy horizontal comparison.

Interestingly, the aggregation weight distributions provided by managers are completely different from the one-hot distributions specified in \btname{}, and there are two distinct trends:
($i$) For SAUE managers in the $1^{\text{st}}$ cross-modal layer, vertically:
textual manager exhibits increasing and then decreasing weights, most favoring $\mathbf{T}_{10}$, unlike $\mathbf{T}_{12}$ and $\mathbf{T}_{7}$ used in \metername{} and \btname{}, respectively;
visual manager exhibits increasing weights, most favoring $\mathbf{V}_{12}$, the same as \metername{} and \btname{}.
($ii$) For AAUE managers in the $2^{\text{nd}}$ to $6^{\text{th}}$ cross-modal layers, horizontally: whether textual or visual managers, they exhibit diverse aggregation weight distributions in different layers.

Overall, comparing the aggregation weight distributions horizontally and vertically, \method{} learns diverse distributions in different cross-modal layers.
This provides strong evidence that the introduced managers can adaptively aggregate uni-modal semantic knowledge for comprehensively cross-modal representation learning.

\begin{figure}[!t]
	\centering
	\includegraphics[width=0.4\textwidth]{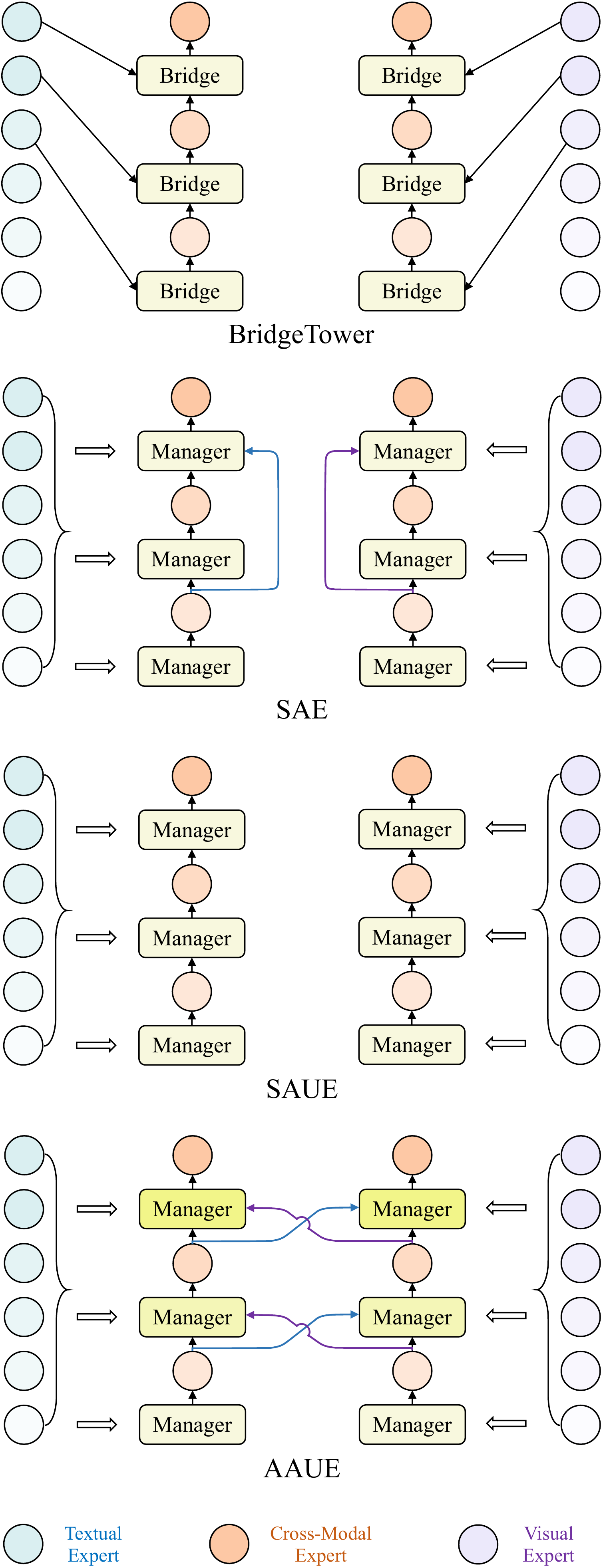}
	\caption{
		Brief illustrations of \btname{} and our \method{} with SAE, SAUE and AAUE managers. 
		Hollow arrows indicate the transmission of multi-layer uni-modal representations in \method{} instead of layer-by-layer transmission in \btname{}. 
		Each uni-modal or cross-modal layer is seen as a uni-modal or cross-modal expert.
		The arrow between the cross-modal expert of the previous layer and the manager of the current layer is to get the cross-modal fused query.
	}
	\label{fig:managers}
\end{figure}

\subsection{Intuitive Comparison Between BT\&MT}
\label{appendix:managers}
We provide brief illustrations in~\Fref{fig:managers} to intuitively compare \btname{} (BT) and \method{} (MT) with different type of managers.

\paragraph{BT~\vs{}~MT with SAUE Managers.}
In~\Tref{tab:number-of-cross-modal-layers}~\&~\ref{tab:different-backbones}, we provide the performance comparison between \btname{} and \method{}.\footnote{
	The re-implemented BridgeTower obtained higher experimental results than the original paper due to the better fine-tuning settings we used for all experiments in~\Sref{sec:investigation-and-analysis}.
}
In fact, \btname{} can be seen as an approximate special case of \method{} with SAUE managers if we replace the learned weights $\mathbf{W}$ in each manager with layer-by-layer one-hot distributions\footnote{
	It means that, for each cross-modal layer, only one uni-modal expert is activated at a time in the bottom-up direction.
} used in \btname{}.
However, as shown in~\Fref{fig:aggregation-weights-saue}, the aggregation weight of textual and visual SAUE managers share a similar progressive trend across cross-modal layers, which is completely different from the distributions in \btname{}.
This allows \method{} with SAUE managers to achieve significant performance gains (from $75.91$ to $76.55$) compared to \btname{}.
Besides, the similar trend of aggregation weights is consistent with the observations in~\Fref{fig:cosine_similarity_cross_uni}, that is, the cosine similarity of aggregated uni-modal representations between managers is always similar to $1$.

\paragraph{SAUE Manager~\vs{}~AAUE Manager.}
When we compare~\Fref{fig:aggregation-weights-pre-trained-aaue}~\&~\ref{fig:aggregation-weights-saue}, their respective aggregation weight distributions are completely different.
This further demonstrates that compared with SAUE managers, AAUE managers can adaptively \textbf{generates} different aggregation weights for different tokens in different samples.
Interestingly, the first column of two figures both comes from the SAUE managers, but the distributions are still clearly different.
We presume that high-layer AAUE managers may help low-layer SAUE managers \textbf{rectify} their management of experts.

We also provide the visualizations of aggregation weights of SAE and AAUE managers without VLP in~\Fref{fig:aggregation-weights-sae}~\&~\ref{fig:aggregation-weights-aaue}. 
Comparing the visualization of three types of managers without VLP, we can find that 
($i$) the learned aggregation weights of SAE and SAUE managers are still a little close to the average initialization we used and they all share a similar progressive trend across cross-modal layers;
($ii$) for each AAUE manager, its generated aggregation weights vary significantly across $6$ uni-modal experts; comparing different cross-modal layers, the distribution of aggregation weights generated by the AAUE manager is also very different.

\section{Related Work}

\paragraph{Vision-Language Models.}
Although VL models differ in model architecture, most of them use uni-modal encoders to extract visual and textual representations, and then fuse them in a cross-modal encoder, which can be unified into the \twotower{} architecture~\citep{lu2019vilbert,su2019vl,chen2020uniter,li2020unicoder,li2020oscar,zhou2020unified,kim2021vilt,radford2021learning,jia2021scaling,li2021align,li2021unimo,li2022blip,dou2021meter,wang2021vlmo,wang2021simvlm,wang2022OFA,wang2022image,yu2022coca}.
As a representative model, \metername{}~\citep{dou2021meter} adopts pre-trained uni-modal encoders and feeds their last-layer representations into the cross-modal encoder.
\btname{}~\citep{xu2022bridge} proposes building layer-by-layer connections between the top uni-modal layers and each cross-modal layer
to utilize different uni-modal layer representations.
However, they still cannot provide adaptive and effective aggregation of multi-layer pre-trained uni-modal representations in each cross-modal layer.

\paragraph{Multi-Layer Representation Aggregation.}
The effectiveness of layer representation aggregation in learning comprehensive representations has been well demonstrated in vision~\citep{lin2017feature,huang2017densely,yu2018deep,xie2021segformer} and language~\citep{peters-etal-2018-deep, wang-etal-2018-multi-layer,wang-etal-2019-learning-deep,wei-etal-2020-multiscale}.
Recent VL models also explore utilization of multi-layer uni-modal representations for better cross-modal representation learning.
\metername{} feeds the weighted sum of uni-modal representations into the first cross-modal layer.
\btname{} introduces bridges into \metername{} so that different uni-modal layer representation are fed layer by layer into each cross-modal layer.
In this work, \method{} explores adaptive and effective aggregation of multi-layer uni-modal representations via well-designed managers.

\section{Conclusion}
We propose \method{}, a novel VL model architecture that gathers and combines the insights of pre-trained uni-modal experts at different levels via the introduced managers in each cross-modal layer.
The feasibility of various designs of managers is well explored, and the effectiveness of \method{} on various downstream VL tasks is well demonstrated.
More comprehensive cross-modal alignment and fusion in each cross-modal layer is achieved by adaptive aggregation of different levels of uni-modal semantic knowledge.
We hope that our work can inspire more research on how to better exploit multi-layer pre-trained uni-modal representations for cross-modal representation learning.

\section*{Limitations}
In this work, we propose managers that allow adaptive aggregation of uni-modal layer representations in each cross-modal layer.
Inevitably, AAUE managers significantly improve performance which slightly increasing the computational budget, as we detailed discussed in \Apref{appendix:computational_budget}.
This needs to be further optimized in the future.
Analysis and optimization are also needed for the other types of managers as shown in \Apref{appendix:other-managers}.
Moreover, as shown in~\Fref{fig:number-of-uni-modal-representations}, the performance of \method{} first increases gradually with the number of uni-modal representations, but then stops increasing and even decreases when the number of uni-modal representations exceeds $6$.
How to obtain better \method{} performance using a lower computational budget while utilizing more insights of uni-modal experts, especially when scaling the model, \eg{} $24$-layer CLIP-ViT L-224/16 and $24$-layer RoBERTa\modellarge{}, is a question worth further exploration.
For example, designing reasonable sparse activation functions for managers in \method{}, instead of simple top-$\mathrm{N}$ or top-$\mathrm{p}$ sampling (which did not work well in our preliminary experiments).

\section*{Acknowledgements}
This work was supported by the National Key R\&D Program of China via grant 2020AAA0106501 and the National Natural Science Foundation of China (NSFC) via grant 62236004 and 61976072.

\bibliography{anthology,custom}
\bibliographystyle{acl_natbib}

\appendix

\section{Implementation Details}
\label{appendix:implementation_details}

\subsection{Vision-Language Pre-training}
We use two commonly used VLP objectives.

\paragraph{Masked Language Modeling.}
For MLM, we follow the conditional masking approach used in UNITER~\citep{chen2020uniter} that randomly masks $15\%$ of the tokens in the text token sequence while keeping the image patch sequence unchanged.
The model is then trained to predict the original masked tokens given the incomplete text sequence and the complete image patch sequence.
The masking strategy and MLM task head we use are the same as RoBERTa. The output top-layer representation of the textual part of the cross-modal encoder is used as input for the MLM task head.

\paragraph{Image-Text Matching.}
For ITM, both matched and mismatched image-text pairs are fed into the model with equal probability. The model is trained to predict whether a given image-text pair is a matched (positive) or a mismatched (negative) pair.
The output top-layer representations of $\texttt{[class]}$ and $\texttt{[<s>]}$ tokens are activated by the non-linear function $\texttt{Tanh}$.
Then the concatenation of the above output representations is fed into a linear classifier with cross-entropy loss for binary classification.

\paragraph{Pre-training Settings.}
\Tref{tab:statistics_pretrain} shows the statistics of the pre-train datasets.
Following previous work~\citep{kim2021vilt,chen2020uniter,li2021align,dou2021meter}, we adopt four public image-caption datasets for pre-training, including Conceptual Captions (CC)~\citep{sharma-etal-2018-conceptual}, SBU Captions (SBU)~\citep{NIPS2011_5dd9db5e}, MSCOCO Captions (COCO)~\citep{chen2015microsoft}, and Visual Genome (VG)~\citep{krishna2017visual}.
The total numbers of the unique images and image-caption pairs in the combined training data are $4$M and $9$M.~\Tref{tab:hyperparam_pretrain} describes the hyperparameters for pre-training the \method{}.
The learning rate of the cross-modal encoder is five times higher than that of uni-modal encoders~\citep{dou2021meter}.

\begin{table}[t]
	\tablestyle{6pt}{1.1}
	\adjustbox{width=0.9\linewidth}{
		\begin{tabular}{l|cccc}
			            & COCO & VG   & CC   & SBU  \\
			\shline
			\#~Images   & 113K & 108K & 2.9M & 860K \\
			\#~Captions & 567K & 4.8M & 2.9M & 860K \\
		\end{tabular}
	}
	\caption{
		Statistics of the pre-train datasets. We remove duplicate image-caption pairs in VG~\citep{kim2021vilt, dou2021meter} and only 2.9M image-caption pairs can be downloaded in CC.
	}
	\label{tab:statistics_pretrain}
\end{table}

\begin{figure*}[!h]
	\centering
	\includegraphics[width=\textwidth]{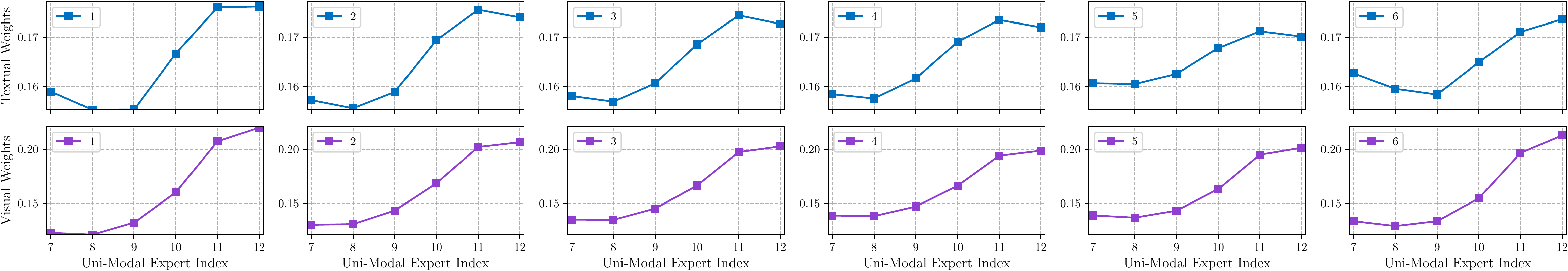}
	\caption{
		A visualization of aggregation weights of textual and visual SAE managers in each cross-modal layer.
		The X-axis is the index of the uni-modal expert, and the legend shows the index of the cross-modal layer.\looseness=-1
	}
	\label{fig:aggregation-weights-sae}
\end{figure*}

\begin{figure*}[!h]
	\centering
	\includegraphics[width=\textwidth]{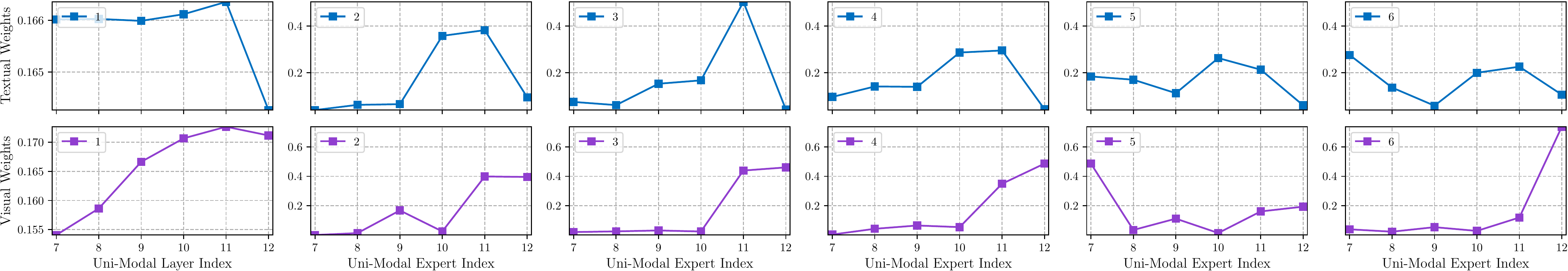}
	\caption{
		A visualization of aggregation weights of textual and visual AAUE managers in each cross-modal layer.
		The X-axis is the index of the uni-modal expert, and the legend shows the index of the cross-modal layer.\looseness=-1
	}
	\label{fig:aggregation-weights-aaue}
\end{figure*}

\begin{table*}[!h]
	\tablestyle{6pt}{1.05}
	\adjustbox{width=0.9\linewidth}{
		\begin{tabular}{rc|cc|cc}
			{Visual}    & {Textual} & \multicolumn{2}{c|}{VQAv2 Test-Dev} & \multicolumn{2}{c}{Flickr30K \rmean{}}                                            \\
			Backbone    & Backbone  & \btname{}                                  & \multicolumn{1}{c|}{\method{}}              & \btname{}     & \multicolumn{1}{c}{\method{}}         \\
			\shline
			DeiT B-224/16 & RoBERTa   & 71.22                               & 72.20 ($\uparrow$\,0.98)               & 87.63 & 88.72($\uparrow$\,1.09)          \\
			ViT B-224/16  & RoBERTa   & 72.82                               & 73.67 ($\uparrow$\,0.85)               & 90.48 & 90.92($\uparrow$\,0.44)          \\
			ViT B-384/16  & RoBERTa   & 72.94                               & 73.80 ($\uparrow$\,0.86)               & 90.51 & 90.96($\uparrow$\,0.45)          \\
			CLIP-ViT B-224/32 & RoBERTa   & 73.73                               & 74.79 ($\uparrow$\,1.06)               & 91.33 & 91.76($\uparrow$\,0.43)          \\
			CLIP-ViT B-224/16 & BERT      & 75.74                               & 76.36 ($\uparrow$\,0.62)               & 92.84 & 93.42($\uparrow$\,0.58)          \\
			CLIP-ViT B-224/16 & RoBERTa   & 75.91                               & \textbf{76.65} ($\uparrow$\,0.74)      & 93.33 & \textbf{93.97}($\uparrow$\,0.64) \\
		\end{tabular}
	}
	\caption{
		Performance of \btname{} and \method{} with different visual and textual backbones.
		B, N and M in ``ViT B-N/M'' denote the model size, image resolution and patch size, respectively.
	}
	\label{tab:different-backbones}
\end{table*}

\subsection{Fine-Tuning on Downstream Tasks}
\paragraph{Dataset Setting.}
Standard settings and splits are used for all datasets.
For Flickr30K dataset~\citep{young-etal-2014-image}, we follow the standard Karpathy Split~\citep{karpathy2015deep}.
For VQAv2~\citep{balanced_vqa_v2} dataset, we follow the common practice~\citep{balanced_vqa_v2, teney2018tips}: convert VQAv2 to a classification task with $3,129$ answer classes; train the model with training data and validation data, and evaluate the model on the Test-Dev and Test-Std data.

\paragraph{Image Augmentation.}
We follow previous works~\citep{li2021align,li2022blip} to use RandomResizedCrop, RandomHorizontalFlip, and RandAugment~\citep{cubuk2020randaugment} to augment the images.

\paragraph{Fine-Tuning Strategy.}
For visual question answering, visual entailment and visual reasoning, the fine-tuning strategy is similar to the strategy we used in ITM.
For image-text retrieval, we follow the approach used in ALBEF~\citep{li2021align} to optimize our model with both image-text contrastive (ITC) and ITM objectives.
In the training phase, we first add two linear projections on top of the uni-modal encoders and calculate the contrastive similarity of uni-modal representations of image-text pairs by dot product to compute the ITC loss.
Formerly, negative image-text pairs in ITM loss are sampled randomly.
However, after computing the ITC loss, we can use contrastive similarity distribution to sample one hard in-batch negative text (image) for each image (text) in a mini-batch.
In the inference phase, we first compute the contrastive similarity for all images and texts, and then select the top-k candidates based on their contrastive similarity. We then calculate their ITM scores for these candidates to determine the final ranking.

\paragraph{Fine-Tuning Settings.}
Similar to the image-text matching (ITM) pre-training objective, we pass the final representation of $\texttt{[class]}$ token and $\texttt{[<s>]}$ token to the non-linear layer activated by $\texttt{Tanh}$, and feed the concatenation of the output into a linear classifier (Flickr30K) or an MLP classifier(VQAv2, SNLI-VE and NLVR$^2$). 
We apply cross-entropy loss for SNLI-VE, NLVR$^2$ and Flickr30K and binary cross-entropy loss for VQAv2~\citep{kim2021vilt,dou2021meter}.
Fine-tuning hyperparameters for VQAv2, SNLI-VE, NLVR$^2$, and Flickr30K are given in~\Tref{tab:hyperparam_finetune}.

\section{Switch Visual and Textual Backbones}
We experiment with different pre-trained visual and textual backbones as uni-modal encoders to further investigate the impact on performance of the managers of \method{} compared to the bridges of \btname{}.
As shown in~\Tref{tab:different-backbones}, regardless of the visual and textual backbones we apply, \method{} significantly and consistently outperforms \btname{} on both datasets.
This further proves the effectiveness and generalization of our proposed \method{} architecture and managers, which can provide adaptive and effective aggregation of multi-layer uni-modal representations for vision-language representation learning.

\begin{table*}[!h]
	\tablestyle{3pt}{1.1}
	\adjustbox{width=\linewidth}{
		\begin{tabular}{lcc|ccc|l|l}
			\multirow{2}{*}{Model}  & {Manager}                & {Manager}                                                        & {\#~Params}             & {\#~FLOPs}              & {Inference Time}            & \multicolumn{1}{c|}{VQAv2}    & \multicolumn{1}{c}{Flickr30K} \\
			                        & \multicolumn{1}{c}{Type} & \multicolumn{1}{c|}{Visual Query}                                & \multicolumn{1}{c}{(M)} & \multicolumn{1}{c}{(G)} & \multicolumn{1}{c|}{(ms)} & \multicolumn{1}{c|}{Test-Dev} & \multicolumn{1}{c}{\rmean{}}  \\
			\shline
			\btname{}\modelbase{} * & -                        & -                                                                & 326.58                  & 101.25                  & 39.43$\pm$1.55              & 75.91                         & 93.33                         \\
			\hdashline
			\method{}\modelbase{}   & SAUE                     & -                                                                & 326.77                  & 101.34                  & 41.12$\pm$1.41              & 76.55 ($\uparrow$\,0.64)      & 93.73 ($\uparrow$\,0.40)      \\
			\method{}\modelbase{}   & AAUE                     & $\mathbf{C}^\mathrm{V}_{\ell-1}$                                 & 326.77                  & 101.35                  & 41.80$\pm$1.05              & 76.52 ($\uparrow$\,0.61)      & 93.84 ($\uparrow$\,0.51)      \\
			\method{}\modelbase{}   & AAUE                     & $\mathbf{C}^\mathrm{V}_{\ell-1}, \mathbf{C}^\mathrm{T}_{\ell-1}$ & 338.64                  & 105.52                  & 43.20$\pm$1.37              & 76.65 ($\uparrow$\,0.74)      & 93.97 ($\uparrow$\,0.64)      \\
		\end{tabular}
	}
	\caption{
		Computational budget and downstream task performance without VLP for \btname{} and \method{}. 
		* denotes our re-implementation.
	}
	\label{tab:computational_budget}
\end{table*}

\section{Computational Budget}
\label{appendix:computational_budget}
Table~\ref{tab:computational_budget} shows the computational budget and downstream task performance without VLP for \btname{} and \method{}, including the number of parameters, the number of FLoating-Point operations (FLOPs)\footnote{We use Facebook Research's \href{https://github.com/facebookresearch/fvcore}{fvcore} to calculate FLOPs.}.
We measure the average inference time of processing 1 VQA instance over 10K runs on 1 NVIDIA TITAN V GPU. The sequence length is $50$, and the image resolution is $384 \times 384$.
Compared with \btname{} ($1^{\text{st}}$ row), \method{} ($4^{\text{th}}$ row) uses an acceptable additional computational budget ($3.69\%$ parameters, $4.22\%$ FLOPs, and $3.77$ms inference time) and achieves significant performance improvements of $0.74\%$ and $3.1\%$ on VQAv2 and Flickr30K, respectively.
We further analyze other well-performed variants of \method{} in the $2^{\text{nd}}$ and $3^{\text{rd}}$ rows.
It is worth noting that the two variants share a similar computational budget as \btname{}, but achieve better performance.
This not only demonstrates the efficiency and effectiveness of our \method{} architecture, but also reminds us that the cross-modal fused query via the cross-attention mechanism is the main reason for the additional computational budget of \method{} ($4^{\text{th}}$ row), as it is the only difference between the $3^{\text{rd}}$ and $4^{\text{th}}$ row models.
This inspires us to explore a more efficient method to fuse $\mathbf{C}^\mathrm{V}_{\ell-1}$ and $\mathbf{C}^\mathrm{T}_{\ell-1}$ to get the cross-modal fused query in the future.

\section{Details on Cross-Attention and Concat-Attention Managers}
\label{appendix:other-managers}

\begin{table*}[t]
	\tablestyle{3pt}{1.1} 
	\adjustbox{width=\linewidth}{
		\begin{tabular}{lrc|cc|cc|cc|cc}
			\multirow{2}{*}{Model}                              & {\#~Pre-train} & {Visual}          & \multicolumn{2}{c|}{VQAv2} & \multicolumn{2}{c|}{SNLI-VE} & \multicolumn{2}{c|}{NLVR$^2$} & \multicolumn{2}{c}{Flickr30K}                                                     \\
			                                                    & Images~        & Backbone          & Test-Dev                   & Test-Std                     & Dev                           & Test                          & Dev        & Test-P     & IR@1       & TR@1       \\
			\shline
			\multicolumn{11}{l}{ { \it{Base-size models pre-trained on 4M public data} } }                                                                                                                                                                                           \\
			\hline
			ViLT\modelbase{}~\citep{kim2021vilt}                & 4M             & ViT B-384/32      & 71.26                      & -                            & -                             & -                             & 75.70      & 76.13      & 64.4       & 83.5       \\
			UNITER\modelbase{}~\citep{chen2020uniter}\,$\ast$   & 4M             & Faster R-CNN      & 72.70                      & 72.91                        & 78.59                         & 78.28                         & 77.18      & 77.85      & 72.52      & 85.90      \\
			VILLA\modelbase{}~\citep{gan2020large}\,$\ast$      & 4M             & Faster R-CNN      & 73.59                      & 73.67                        & 79.47                         & 79.03                         & 78.39      & 79.30      & 74.74      & 86.60      \\
			UNIMO\modelbase{}~\citep{li2021unimo}               & 4M             & Faster R-CNN      & 73.79                      & 74.02                        & 80.00                         & 79.10                         & -          & -          & 74.66      & 89.70      \\
			ALBEF\modelbase{}~\citep{li2021align}\,$\ast$       & 4M             & DeiT B-224/16     & 74.54                      & 74.70                        & 80.14                         & 80.30                         & 80.24      & 80.50      & 82.8       & 94.3       \\
			VinVL\modelbase{}~\citep{zhang2021vinvl}            & 5.7M           & ResNeXt-152       & 75.95                      & 76.12                        & -                             & -                             & 82.05      & 83.08      & -          & -          \\
			\metername{}-Swin\modelbase{}~\citep{dou2021meter}  & 4M             & Swin B-384/32     & 76.43                      & 76.42                        & 80.61                         & 80.45                         & 82.23      & 82.47      & 79.02      & 92.40      \\
			\vlmoname{}\modelbase{}~\citep{wang2021vlmo}        & 4M             & BEiT B-224/16     & 76.64                      & 76.89                        & -                             & -                             & 82.77      & 83.34      & 79.3       & 92.3       \\
			\metername{}-CLIP\modelbase{}~\citep{dou2021meter}  & 4M             & CLIP-ViT B-224/16 & 77.68                      & 77.64                        & 80.86                         & 81.19                         & 82.33      & 83.05      & 82.22      & 94.30      \\
			\btname{}\modelbase{}~\citep{xu2022bridge}          & 4M             & CLIP-ViT B-224/16 & 78.66                      & 78.73                        & 81.11                         & 81.19                         & 81.85      & 83.09      & 85.83      & 94.73      \\
			\method{}\modelbase{}~(\bf{Ours})                   & 4M             & CLIP-ViT B-224/16 & \bf{79.39}                 & \bf{79.15}                   & \bf{81.26}                    & \bf{81.44}                    & \bf{82.81} & \bf{83.34} & \bf{86.56} & \bf{95.64} \\
			\hline
			\multicolumn{11}{l}{ { \it{Models pre-trained on more data and/or with larger size}}}                                                                                                                                                                                    \\
			\hline
			UNITER\modellarge{}~\citep{chen2020uniter}\,$\ast$  & 4M             & Faster R-CNN      & 73.82                      & 74.02                        & 79.39                         & 79.38                         & 79.12      & 79.98      & 75.56      & 87.30      \\
			VILLA\modellarge{}~\citep{gan2020large}\,$\ast$     & 4M             & Faster R-CNN      & 74.69                      & 74.87                        & 80.18                         & 80.02                         & 79.76      & 81.47      & 76.26      & 87.90      \\
			UNIMO\modellarge{}~\citep{li2021unimo}              & 4M             & Faster R-CNN      & 75.06                      & 75.27                        & 81.11                         & 80.63                         & -          & -          & 78.04      & 89.40      \\
			ALBEF\modelbase{}~\citep{li2021align}\,$\ast$       & 14M            & DeiT B-224/16     & 75.84                      & 76.04                        & 80.80                         & 80.91                         & 82.55      & 83.14      & 85.6       & 95.9       \\
			VinVL\modellarge{}~\citep{zhang2021vinvl}           & 5.7M           & ResNeXt-152       & 76.52                      & 76.63                        & -                             & -                             & 82.67      & 83.98      & -          & -          \\
			BLIP\modelbase{}~\citep{li2022blip}\,$\ast$         & 14M            & DeiT B-224/16     & 77.54                      & 77.62                        & -                             & -                             & 82.67      & 82.30      & 87.2       & 96.6       \\
			SimVLM\modelbase{}~\citep{wang2021simvlm}\,$\star$  & 1.8B           & ResNet-101        & 77.87                      & 78.14                        & 84.20                         & 84.15                         & 81.72      & 81.77      & -          & -          \\
			BLIP\modelbase{}~\citep{li2022blip}\,$\ast$         & 129M           & DeiT B-224/16     & 78.24                      & 78.17                        & -                             & -                             & 82.48      & 83.08      & 87.3       & 97.3       \\
			SimVLM\modellarge{}~\citep{wang2021simvlm}\,$\star$ & 1.8B           & ResNet-152        & 79.32                      & 79.56                        & 85.68                         & 85.62                         & 84.13      & 84.84      & -          & -          \\
			\vlmoname{}\modellarge{}~\citep{wang2021vlmo}       & 4M             & BEiT L-224/16     & 79.94                      & 79.98                        & -                             & -                             & 85.64      & 86.86      & 84.5       & 95.3       \\
			SimVLM\modelhuge{}~\citep{wang2021simvlm}\,$\star$  & 1.8B           & Larger ResNet-152 & 80.03                      & 80.34                        & 86.21                         & 86.32                         & 84.53      & 85.15      & -          & -
		\end{tabular}
	}
	\caption{
		Comparisons with previous models on various downstream VL tasks.
		The best score is bolded. 
		B, N and M in ``ViT B-N/M'' denote the model size, image resolution and patch size, respectively.
		$\ast$ indicates that the model also uses VG-QA data to fine-tune on VQAv2.
		$\star$ denotes the model is trained from scratch. ``\# Pre-train Images'' denotes the number of unique images used in VLP.
	}
	\label{tab:detailed-main-results}
\end{table*}

\paragraph{Cross-Attention Managers.}
We implement the standard cross-attention mechanism~\citep{vaswani2017attention} and reduce the linear projection layer for value to save computational budget.\footnote{The calculation of cross-modal fused query also uses this simplified version of the cross-attention mechanism.}
Take the visual manager for example, it takes $\mathbf{C}^\mathrm{V}_{\ell-1} \in \mathbb{R}^{\mathrm{L} \times \mathrm{D}}$ as the query, and the first token of multi-layer uni-modal representations, \ie{} $\mathbf{V}[:, 0] \in \mathbb{R}^{\mathrm{N} \times \mathrm{D}}$, as the key. Hence, the shape of generated aggregation weights is $\mathrm{N} \times \mathrm{L}$, which can be broadcast to the aggregation weights $\mathbf{W}_\mathrm{A}\!\in\!\mathbb{R}^{\mathrm{N} \times \mathrm{L} \times \mathrm{D}}$. 
The following calculation is the same as AAUE managers in~\Fref{fig:details}. 
The results in~\Tref{tab:type-and-query} show a significant decrease compared to other managers on Flickr30K. We leave the detailed analysis of this phenomenon to the future work.

\paragraph{Concat-Attention Managers.}
Take the visual manager as an example, it broadcasts $\mathbf{C}^\mathrm{V}_{\ell-1}\!\in\!\mathbb{R}^{\mathrm{L} \times \mathrm{D}}$ to $\mathbb{R}^{\mathrm{N} \times \mathrm{L} \times \mathrm{D}}$, and concatenates it with $\mathbf{V}\!\in\!\mathbb{R}^{\mathrm{N} \times \mathrm{L} \times \mathrm{D}}$ along the last dimension as the concatenated query.
It then directly projects the query to $\mathbf{W}_\mathrm{A}\!\in\!\mathbb{R}^{\mathrm{N} \times \mathrm{L} \times \mathrm{D}}$. The following calculation is the same as AAUE managers in~\Fref{fig:details}.
In fact, this type of manager is different from all other managers from the perspectives of the generated aggregation weights.
Although its aggregation weights delve into the feature dimension of $\mathbf{C}^\mathrm{V}_{\ell-1}$ and $\mathbf{V}$, the substantially increased number of parameters and computational cost do not result in a significant performance gain, making it impractical and inefficient.
More efficient variants of this type of manager should be investigated in the future.

\section{Detailed Comparison with Previous Arts}
Due to the space limitations, we omit some baselines and details in~\Tref{tab:main-results}. 
Here we provide more details on the comparison with previous arts in~\Tref{tab:detailed-main-results}.

\begin{table*}[t]
	\tablestyle{6pt}{1.1}
	\adjustbox{width=0.47\linewidth}{
		\begin{tabular}{l|cc}
			Hyperparameters            & \method{}               \\
			\shline
			Number of Layers           & $6$                     \\
			Hidden size                & $768$                   \\
			FFN inner hidden size      & $3,072$                 \\
			Number of Attention heads  & $12$                    \\
			Dropout Ratio              & $0.1$                   \\
			Attention dropout          & $0.1$                   \\
			\hline
			Total Steps                & $100$k                  \\
			Batch Size                 & $4,096$                 \\
			Optimizer                  & $\operatorname{AdamW}$  \\
			Learning Rate              & $1e^{-5}$               \\
			Learning Rate Decay        & $\operatorname{Linear}$ \\
			Weight Decay               & $0.01$                  \\
			Warmup Steps               & $10$k                   \\
			Adam $\epsilon$            & $1e^{-8}$               \\
			Adam $\beta_1$             & $0.9$                   \\
			Adam $\beta_2$             & $0.98$                  \\
			\hline
			Center-Crop                & \cmark                  \\
			Random Resized Crop        & \xmark                  \\
			Random Augmentation        & \xmark                  \\
			Random Horizontal Flipping & \xmark                  \\
			\hline
			Textual Encoder            & RoBERTa\modelbase{}     \\
			Visual Encoder             & CLIP-ViT B-224/16              \\
			Patch Size                 & $16$                    \\
			Image Resolution for VLP   & $288$                   \\
		\end{tabular}
	}
	\caption{
		Hyperparameters for pre-training. The first block is the hyperparameters for the cross-modal encoder.
	}
	\label{tab:hyperparam_pretrain}
\end{table*}

\begin{table*}[!h]
	\tablestyle{4pt}{1.1}
	\adjustbox{width=\linewidth}{
		\begin{tabular}{l|cccc}
			Hyperparameters            & VQAv2                   & SNLI-VE                 & NLVR$^2$                & Flickr30K               \\
			\shline
			Total Epochs               & $10$                    & $4$                     & $5$                     & $20$                    \\
			Batch Size                 & $576$                   & $64$                    & $256$                   & $512$                   \\
			Optimizer                  & $\operatorname{AdamW}$  & $\operatorname{AdamW}$  & $\operatorname{AdamW}$  & $\operatorname{AdamW}$  \\
			Learning Rate              & $9e^{-6}$               & $3e^{-6}$               & $1.4e^{-5}$             & $6e^{-6}$               \\
			Learning Rate Decay        & $\operatorname{Linear}$ & $\operatorname{Linear}$ & $\operatorname{Linear}$ & $\operatorname{Linear}$ \\
			Weight Decay               & $0.06$                  & $0.01$                  & $0.01$                  & $0.01$                  \\
			Warmup Ratio               & $0.06$                  & $0.06$                  & $0.1$                   & $0.1$                   \\
			Adam $\epsilon$            & $1e^{-8}$               & $1e^{-8}$               & $1e^{-8}$               & $1e^{-8}$               \\
			Adam $\beta_1$             & $0.9$                   & $0.9$                   & $0.9$                   & $0.9$                   \\
			Adam $\beta_2$             & $0.98$                  & $0.98$                  & $0.98$                  & $0.98$                  \\
			\hline
			Center-Crop                & \xmark                  & \xmark                  & \xmark                  & \xmark                  \\
			Random Resized Crop        & \cmark                  & \cmark                  & \cmark                  & \cmark                  \\
			Random Augmentation        & \cmark                  & \cmark                  & \cmark                  & \cmark                  \\
			Random Horizontal Flipping & \xmark                  & \cmark                  & \cmark                  & \cmark                  \\
			\hline
			Textual Encoder            & RoBERTa\modelbase{}     & RoBERTa\modelbase{}     & RoBERTa\modelbase{}     & RoBERTa\modelbase{}     \\
			Visual Encoder             & CLIP-ViT B-224/16              & CLIP-ViT B-224/16              & CLIP-ViT B-224/16              & CLIP-ViT B-224/16              \\
			Patch Size                 & $16$                    & $16$                    & $16$                    & $16$                    \\
			Image Resolution for FT    & $576$                   & $384$                   & $384$                   & $384$                   \\
			Loss Function              & BCE                     & CE                      & CE                      & CE                      \\
		\end{tabular}
	}
	\caption{
		Hyperparameters for fine-tuning \method{} on various downstream VL tasks. FT denotes fine-tuning. CE and BCE are short for cross-entropy loss and binary cross-entropy loss, respectively.
	}
	\label{tab:hyperparam_finetune}
\end{table*}

\end{document}